\documentclass[journal,transmag]{IEEEtran}
\usepackage{cite}
\usepackage{amsmath,amssymb,amsfonts}
\usepackage{amsthm}
\usepackage{algorithmic}
\usepackage{graphicx}
\usepackage{textcomp}
\usepackage{cleveref}
\usepackage{enumerate}
\usepackage{booktabs}
\usepackage{tabularx}
\usepackage{url}
\usepackage{multirow}
\usepackage{subfiles}
\usepackage{booktabs}\usepackage{nicefrac}
\usepackage[linesnumbered,ruled]{algorithm2e}
\usepackage{xcolor}

\newcommand{\citep}[1]{\cite{#1}}
\renewcommand{\Cref}[1]{\cref{#1}}

\usepackage[colorinlistoftodos,textsize=footnotesize]{todonotes}

\newcommand\vghl[1]{\color{red}}
\newcommand\hao[1]{\color{blue}}

\usepackage{mathtools}

\graphicspath{{images/}}

\newcommand{\mypar}[1]{\vskip 1em\noindent \textbf{#1.} }

\allowdisplaybreaks

\date{}
\begin{document}
%


\markboth{Preprint}
{Chen\MakeLowercase{\textit{et al.}}: Gap filling thin structure}

\title{Blind Inpainting of Large-scale Masks of Thin Structures with Adversarial and Reinforcement Learning}

\author{Hao Chen, Mario Valerio Giuffrida, Peter Doerner, and Sotirios A. Tsaftaris, \IEEEmembership{Senior Member, IEEE}

\thanks{Paper submitted on 29th November 2019.}

\thanks{MV. Giuffrida is with the School of Computing, Edinburgh Napier University, 10 Colinton Road, EH10 5DT, Edinburgh, UK (v.giuffrida@napier.ac.uk).}
\thanks{P. Doerner is with the School of Biological Sciences, Daniel Rutherford Building, University of Edinburgh, EH9 3BF, Edinburgh, UK (Peter.Doerner@ed.ac.uk).}
\thanks{S. A. Tsaftaris is with the Institute for Digital Communications, School of Engineering, University of Edinburgh Thomas Bayes Road, EH9 3FG, Edinburgh, UK, and with The Alan Turing Institute, 96 Euston Road, NW1 2DB, London, UK (s.tsaftaris@ed.ac.uk).}}


%



\maketitle
\begin{abstract}
Several imaging applications (vessels, retina, plant roots, road networks from satellites) require the accurate segmentation of thin structures for subsequent analysis. Discontinuities (gaps) in the extracted foreground may hinder down-stream image-based analysis of biomarkers, organ structure and topology. In this paper, we propose a general post-processing technique to recover such gaps in large-scale segmentation masks.  We cast this problem as a blind inpainting task, where the regions of missing lines in the segmentation masks are not known to the algorithm, which we solve with an adversarially trained neural network. One challenge of using large images is the memory capacity of current GPUs. The typical approach of dividing a large image into smaller patches to train the network does not guarantee global coherence of the reconstructed image that preserves structure and topology.  We use adversarial training and reinforcement learning (Policy Gradient) to endow the model with both global context and local details. We evaluate our method in several datasets in medical imaging, plant science, and remote sensing. Our experiments demonstrate that our model produces the most realistic and complete inpainted results, outperforming other approaches. In a dedicated study on plant roots we find that our approach is also comparable to human performance. Implementation available at \url{https://github.com/Hhhhhhhhhhao/Thin-Structure-Inpainting}.
\end{abstract}
\begin{IEEEkeywords}
Gap filling, blind inpainting, reinforcement learning, adversarial networks, deep learning.
\end{IEEEkeywords}


\IEEEdisplaynontitleabstractindextext

%
\IEEEpeerreviewmaketitle

\section{Introduction}
\label{sec:introduction}

\begin{figure}
    \centering
    \includegraphics[width=\linewidth]{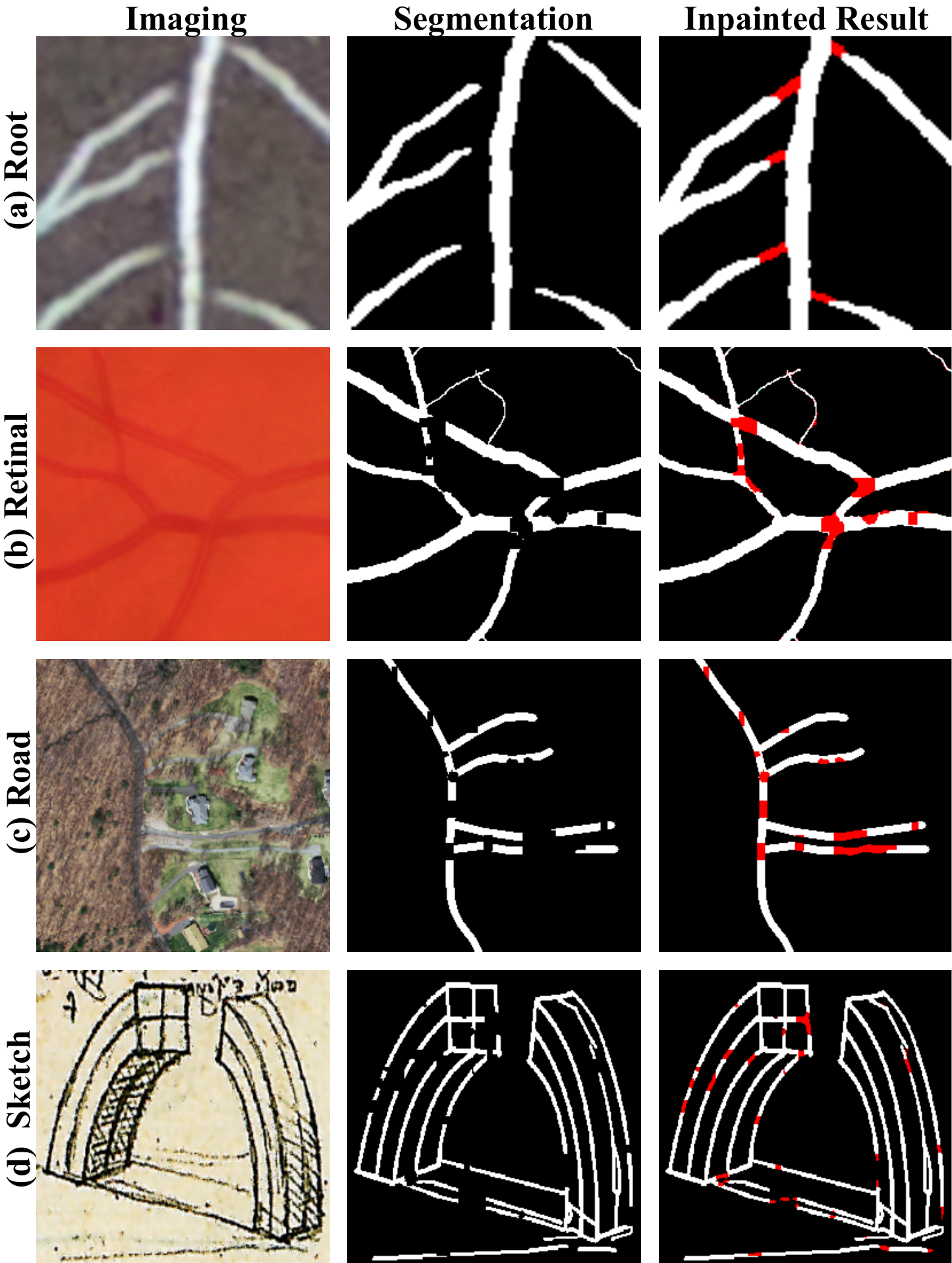}
    \caption{Thin structure inpainting examples of (a) plant roots; (b) retinal vessel; (c) roads; (d) line sketch. Part of the objects is missing in the segmentation (middle column), either because of occlusion  or under-segmentation (in (a) gaps are real; in (b)-(d) artificially introduced).  Our algorithm (last column) detects and inpaints the gaps automatically (shown in red).}
    \label{fig:fig1-root-inpainting}
    \vspace{-1em}
\end{figure}

For several image-based applications, segmentation occurs as a fundamental step for down-stream analysis.  However, segmentation masks may exhibit discontinuities, due to limitations of sensing resolution, occlusions, or the segmentation algorithm \cite{sekou2019patch, zhang2019aerial}. These discontinuities (gaps) are particularly problematic when imaging (and segmenting) thin structures, such as retinal vessels, nadir images of roads, line drawing sketches or plant roots (c.f. \Cref{fig:fig1-root-inpainting}).

\begin{figure}
    \centering
    \includegraphics[width=0.9\linewidth]{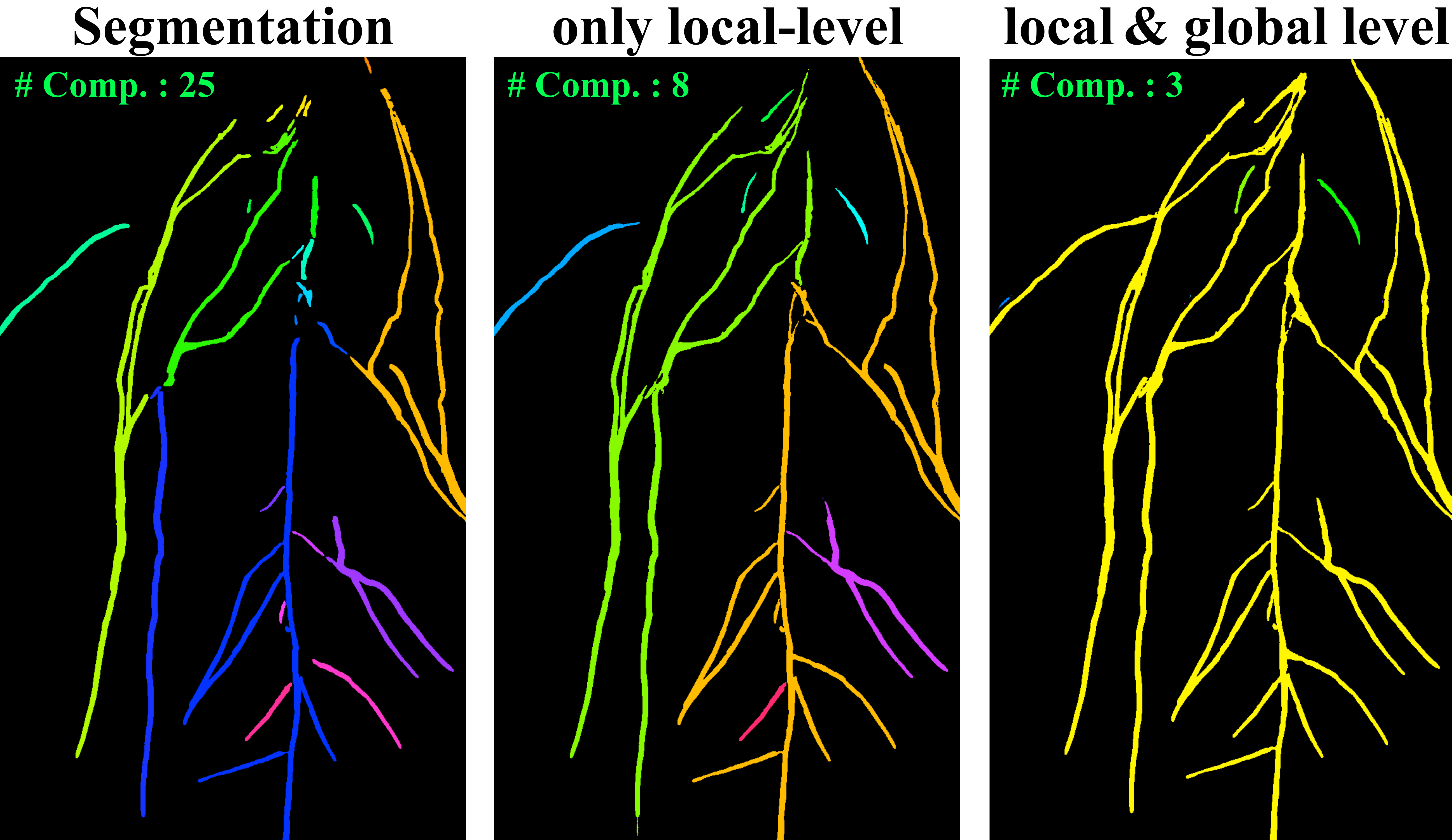}
    \caption{Importance of patch and global context. \textit{Left:} input image; \textit{Middle:} using only the patch-level inpainting; \textit{Right:} using both patch-  and global-level inpainting. We perform connected component labelling and color each component to show the completeness of the root in this image. Clearly, combining both local inpainting with global context considerably improves performance (shown as a reduced number of connected components [\#Comp.]).} 
    \label{fig:compare-gan-gl}
\end{figure}

Here we address the problem of recovering gaps in binary segmentation masks as an inpainting task with deep neural networks. Different from others, we address two main challenges: (i) \textit{blind} inpainting; and (ii) large-scale learning and inference. Specifically, most of the inpainting approaches for natural images are \textit{non-blind}, as the  region to inpaint is known to the algorithm \cite{Pathak2016ContextEF, Iizuka2017GloballyAL, partialconv2017, Yu2018FreeFormII}. In our case, the inpainting is \textit{blind}, because there is no distinction between background and a gap (see middle column in \Cref{fig:fig1-root-inpainting}). Therefore, in real segmentation masks, we do not know where the gaps occur since they appear as background. Another challenge arises from the large-scale size of the segmentation masks in practical applications. For example, retinal vessel segmentation masks in \textit{High-Resolution Fundus} (HRF) \cite{odstrcilik2013retinal} datasets have dimensions $\approx2300 \times 3500$. Such large-scale images cannot be passed on the GPU for training due to GPU memory limits. Down-scaling is not an applicable solution, as it would introduce other gaps due to down-sampling. Most projects break down the large images training the model with smaller patches \cite{7161344, 2016ISPAnIII3473M, maggiori2017convolutional}. However, this does not guarantee global coherence and preservation of global structure, as patches offer a (myopic) field of view. An example is shown in \Cref{fig:compare-gan-gl}: when our model is trained only on patch-level, inpainting quality of the gap is affected, compared to when we train our model with large-scale images.



We propose an adversarial inpainting method for thin structure segmentation masks that combines the advantages of patch- and image-level processing.\footnote{This  manuscript  extends  \cite{Chen2019}: we offer additional experiments in several new datasets in medical imaging, remote sensing, and line sketch datasets; we include comparisons with expert annotators; and we provide detailed ablation studies highlighting the importance of each component in the model.} Our approach starts with large-scale images and extracts at randomly patches and introduces random artificial gaps. A generator network $G$, inpaints at patch level and is trained using a supervised loss comparing inpainted output and input (before the gaps).  A local patch-level discriminator $D_L$, learns a data-driven loss further improving the generator's performance. 
Our key contribution is a global discriminator $D_G$, which is used to assess global image-level coherence and structure preservation. $D_G$ learns to compare image-level inpainting results and original input (without gaps). We use \textit{Policy Gradient} \cite{Sutton1999PolicyGM} --a reinforcement learning technique-- since the processes of random sampling, binarising, and re-composing an inpainted large-scale image (from patches) are random and  non-differentiable.

Our contributions are summarized as follows:
\begin{itemize}

\item A general method for post-processing to correct gaps in binary images of thin structures. To the best of our knowledge this has not been previously addressed. 
\item We perform blind inpainting on large-scale images, where no knowledge of the positions of the regions to inpaint is provided to the network.
	
\item We utilize Policy Gradient to enable the interaction between a global discriminator $D_G$ and the generator $G$ in the presence of a non-differentiable process, while working around GPUs issues with large-scale images.



\item To use real datasets (segmentation masks) that contain gaps we combine synthetic and real data. To bridge the domain gap we introduce a variant of our method replacing the supervised loss with a mask loss  \cite{wu2017maskedloss} and an alternating optimisation strategy. This helps the network to ignore gaps already present in real data whilst reducing data bias and domain shift.


\item We validate our approach in several domains: plant biology (images of roots\footnote{Our images are obtained from an affordable imaging apparatus \cite{Bontpart573139} that captures the chickpea root as it grows. The hope is that via the analysis of these images we can offer tools for breeders to develop more drought-resistant chickpea plants. It is a collaboration with biologists and plant scientists in the UK and Ethiopia. More information is available at \url{http://chickpearoots.org/}.}); medical imaging (retina vessels); remote sensing (images of roads); and art (line sketches). 

\item Our approach outperforms approaches that can be adopted in our context, for example \cite{Sasaki2017JointGD}. For the plant root images we show that our model performs as good as expert annotations.
\end{itemize}

We proceed as follows. In \Cref{sec:rw}, we discuss literature in inpainting. Then, in \Cref{sec:mb} we offer mathematical background on adversarial networks and policy gradient. We detail our approach in \Cref{sec:pm}. Experimental results are shown in \Cref{sec:er}. Finally, \Cref{sec:conclusion} offers the conclusions.

\section{Related Work}
\label{sec:rw}

In the context of natural images, knowing the region to inpaint can be a fair assumption and has been profusely utilised by most of the modern inpainting methods. However, with binary images, deteriorated regions and background appear the same, as an image can only contain either `0' or `1' (c.f. \cref{fig:fig1-root-inpainting}). This is in fact the case of thin structures, which is the focus of this paper. Repairing segmentation masks can thus be seen as a \textit{blind} inpainting problem, where the region to be repaired is unknown. In the following, we first focus our review on thin structure recovery. For completeness, we also briefly discuss deep learning approaches for natural image inpainting.

\subsection{Thin Structure Image Recovery}
The recovery of thin structures in binary segmentation masks is relatively understudied; only recently methods for images of line sketches \cite{Sasaki2017JointGD} and plant roots \cite{Chen2018RootGC,Chen2019} have appeared. 
The authors in \citep{Sasaki2017JointGD} demonstrated that line drawings have enough structure to allow the model to automatically detect and inpaint the gaps without the need for masks indicating the missing regions. This inspired us in conducting blind inpainting for thin-structure segmentation masks of plant roots \cite{Chen2018RootGC}. We later extended this approach by considering local and global discriminators in an attempt to offer large-scale inpainting alleviating GPU memory limitations\cite{Chen2019}. 

This paper builds upon these aiming to offer an approach that has broader applicability demonstrated with additional experiments, a loss that improves  training when data may have inherent gaps (such as roots) \cite{wu2017maskedloss}, and several evaluation and ablation studies including human evaluation (for plant roots). In our experiments we compare with \cite{Sasaki2017JointGD} and  \cite{Chen2018RootGC}.



\subsection{Deep Learning Inpainting for Natural Images}
Inpainting, as proposed in \cite{Bertalmo2000ImageI}, aims to recover a certain deteriorated area $\Omega$ in images. Traditional approaches are PDE-based diffusion methods (inspired by \cite{Bertalmo2000ImageI}), or dictionary-based methods, filling $\Omega$ optimizing a similarity measure over a local manifold of patches (e.g. \cite{Barnes2009PatchMatchAR}). These methods have shown limited performance in recovering complex details, since they rely on relatively low-level image statistics to conduct inpainting with somewhat inferior semantic understanding \cite{Iizuka2017GloballyAL}. With the advent of deep learning, several inpainting architectures have recently been proposed. 

The \textit{Context Encoder} \cite{Pathak2016ContextEF} is one of the first deep learning inpainting methods with an encoder-decoder architecture using both mean squared error (MSE) and adversarial losses for a better inpainting result. To reduce information loss during the down-sampling of the encoder and enlarge the receptive filed of the model, classical convolutions have been replaced with dilated convolutions \cite{Iizuka2017GloballyAL}. Later, partial \cite{partialconv2017} and gated \cite{Yu2018FreeFormII} convolutions have been proposed to evolve the mask channel of gaps along with convolutional operations, preserving the reconstruction details more precisely. Recognizing that contours (edges) provide structural cues, several approaches aim to learn to inpaint natural images aided by the corresponding region boundaries, using edge and contour detection and completion globally \cite{nazeri2019edgeconnect} or locally \cite{Xiong_2019_CVPR} and in multiple-scales \cite{DFNet2019}.

While these projects are inspiring they are for natural images and are not blind as they assume known location of gaps. In our model the gaps are not distinguishable from background. Also, due to patch-based training, they are not able to explicitly capture a global view of a large-scale input image. Instead our method does capture global structure of large-scale inputs while the training is still at patch-level.

\section{Background on policy gradient}
\label{sec:mb}
\label{sec:pg}
Our goal is to use large images as input. To train, we extract patches from an input image (which we assume to be complete without gaps), add gaps artificially, inpaint, binarise, and re-compose a large-scale image. To overcome the non-differentiable processes of sampling and re-composition, motivated by \cite{Dai2017TowardsDA}, we use \textit{Policy Gradient} (PG) \citep{Sutton1999PolicyGM}: a reinforcement learning approach for function approximation. Here, an agent is in a certain state $s \in \mathbb{S}$, interacting with an unknown environment. When an action $a \in \mathbb{A}$ is undertaken by the agent, the environment provides a reward  $r \in \mathbb{R}$. Each state is related to a value function $V^{\pi}(s)$ predicting the expected amount of future rewards the agent can receive in this state by performing a policy $\pi(s)$, which tell the agent what action $a$ to take in the state $s$. The final goal is to learn an optimal policy that maximizes the reward received from the environment. 

\begin{figure*}
    \centering
    \includegraphics[width=\linewidth]{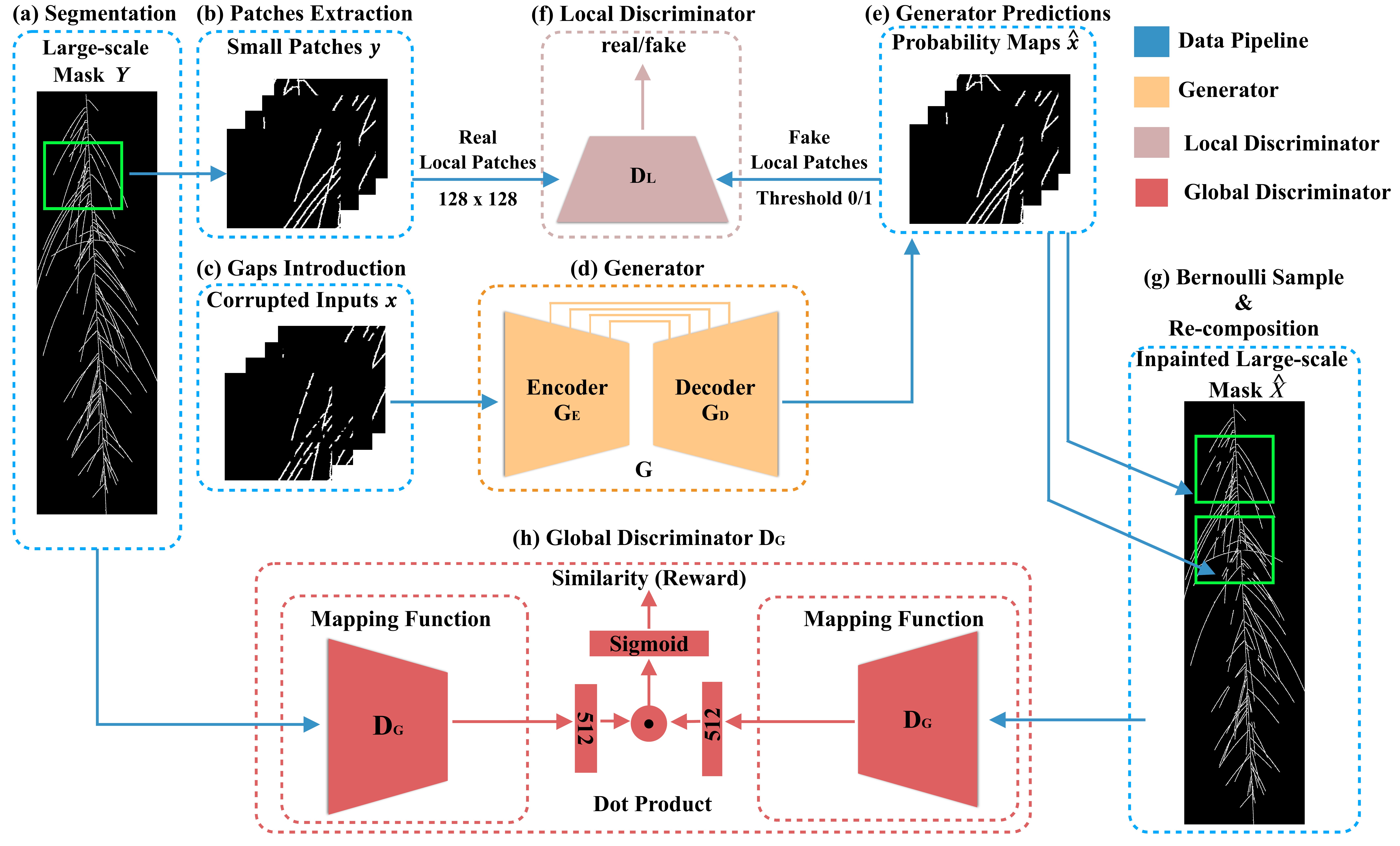}
    \caption{Overview of model architecture and training. (a) A large-scale complete segmentation mask $Y$; (b) Non-overlapping patches $y$ of size $256 \times 256$ are extracted from $Y$; (c) Random gaps are introduced into $y$ to produce corrupted inputs $x$. (d) Inpainting generator network $G$, consisting of an encoder $G_E$ and a decoder $G_D$ with skip connections in between. (e) The prediction of the probability maps $\hat{x}$ obtained by $G$. (f) Local discriminator $D_L$ classifies inpainted $\hat{x}$ and complete patches $y$ aggregating decisions over smaller $128 \times 128$ local patches of the inputs. (g) Bernoulli sampling is conducted on $\hat{x}$ to binarise the inpainted results, which are then used to re-compose the segmentation mask where they were extracted from to obtain an inpainted $\hat{X}$. (h) Global discriminator $D_G$ computes the similarity score between $Y$ and $\hat{X}$.}
    \label{fig:fig7-overview}
\end{figure*}

In the Policy Gradient method, the policy-maker is a function parameterized by $\theta$, $\pi(a|s; \theta)=p(a|s; \theta)$ which predicts the probability of the next action. The reward function depending on this policy is defined as:
\begin{equation}
	\begin{split}
		\mathcal{L}_{PG}(\theta) = & \sum_{s \in \mathbb{S}} d^{\pi}(s) V^{\pi}(s) \\
		= &\sum_{s \in \mathbb{S}} d^{\pi}(s) \sum_{a \in \mathbb{A}} \pi(a|s, \theta) Q^{\pi}(s, a),
	\end{split}
\label{eq:pgorg}
\end{equation}
\noindent where $d^{\pi}(s)$ is the stationary distribution of Markov decision chain for states, and $Q^{\pi}(s,a)$ computes the reward value for a given pair of state and action. However, computing $\nabla_{\theta} \mathcal{L}_{PG}(\theta)$ is typically unfeasible, as it depends on both action selection and the stationary distribution of the states. The policy gradient theorem \citep{Sutton1999PolicyGM} simplifies the computation of the derivation by reformulating \cref{eq:pgorg} as follows:
\begin{equation}
\nabla_{\theta} \mathcal{L}_{PG}(\theta) \propto \mathbb{E}_{s \sim d^{\pi}, a \sim \pi}[Q^{\pi}(s,a)  \nabla_{\theta} \log{\pi(a|s, \theta)}].
\label{eq:policygradienttheorem}
\end{equation}

In this paper, we use REINFORCE \citep{williams1992reinforcement} as policy gradient algorithm. It relies on an estimation of Monte Carlo methods which uses episode samples to update the policy parameter $\theta$: 
\begin{equation}
\nabla_{\theta} \mathcal{L}_{PG}(\theta) = \mathbb{E}_{s \sim d^{\pi}, a \sim \pi}[Q^{\pi}(s_t,a_t) \nabla_{\theta} \log{\pi(a_t|s_t; \theta)}]. 
\label{eq:reinforce}
\end{equation}




\section{Proposed Method}
\label{sec:pm}

We cast the blind recovery of gaps in segmentation masks of thin structures, as an inpainting task using adversarial learning inspired by Conditional Generative Adversarial Networks (cGANs) --an extension of GANs-- \cite{Goodfellow2014GenerativeAN}, which use  
 images as an input (instead of only noise) \cite{Isola2017ImagetoImageTW, CycleGAN2017}.
 Our approach consists of several steps (including given an input image, sampling patches, adding gaps, inpainting, and re-composing an inpainted image), illustrated in \Cref{fig:fig7-overview}, and uses three key networks (detailed network structures in the supplemental): (i) a generator $G$; (ii) a local discriminator $D_L$; and (iii) a global discriminator $D_G$. The generator takes a corrupted patch as input and provides an inpainted version with the gaps filled. The local discriminator assesses if the gaps are correctly filled at the local level. The global discriminator uses the entire image (after re-composition) and helps the generator to assess inpainting quality globally. 

\subsection{Generating Random Gaps}
\label{sec:randomgaps}
\begin{figure}[t]
    \centering
    \includegraphics[width=\linewidth]{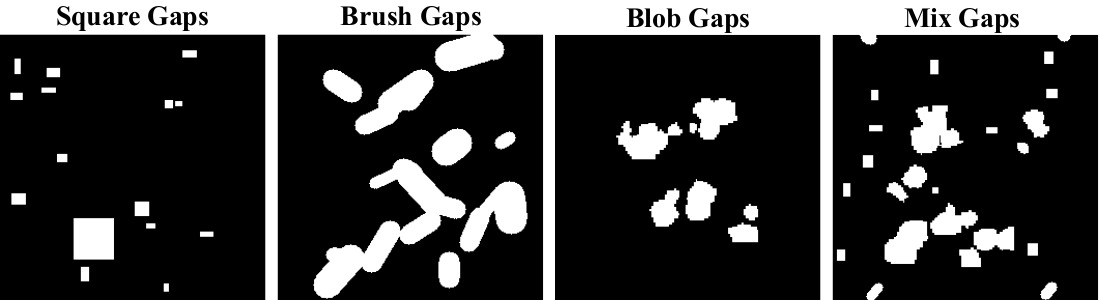}
    \caption{Examples of the artificially generated gaps used during training. Mix gaps refers to the combination of square, brush, and blob gaps.}
    \label{fig:fig4-gaps}
\end{figure}

To train the model, we create corrupted patches, $x$, with artificially introduced random gaps from patches ($y$) extracted from (large-scale) training images $Y$. Typically square gaps are used \cite{Sasaki2017JointGD}. However, square gaps are highly structured, which can lead to overfitting. Instead we introduce gaps of diverse shapes, as \cref{fig:fig4-gaps} shows. We introduce \textit{brush gaps} using the algorithm in \cite{Yu2018FreeFormII}.  However, these gaps do not necessarily fall on foreground pixels and gap edges have structure (i.e. vertical and circular). To further increase diversity, we introduce \textit{blob gaps} \citep{dupont2018probabilistic}, which are even less structured. Since, these gaps are created at pixel-level, to reduce computational overhead during training, we create a library of blobs of size $32 \times 32$. When adding \textit{blob gaps}, we randomly sample from the blob masks and re-scale them to different sizes (accordingly). 

\subsection{Generator $G$}


$G$ takes as input corrupted patches, $x$, ($256 \times 256$) and predicts a probability map, $\hat{x}$, indicating a pixel's likelihood to be foreground. Note that, since our inputs are binary, gap or background pixels are the same (i.e., both are 0 to the network). We do not provide the network the location of gaps. For these reasons, the network is \textit{never} aware where a gap is and thus must jointly learn to detect and inpaint gaps. $G$ is a fully convolutional network \citep{Shelhamer2017fcn} consisting of an encoder $G_E$ and a decoder $G_D$ with skip connections as a U-Net \cite{RFB15unet} (detailed architecture in the supplemental).



To train $G$ we use a supervised cross-entropy loss between $\hat{x}$ predicted by $G$ and patches $y$
\begin{equation} 
\mathcal{L}_{CE} = \mathbb{E}_{y \sim p_{data}, x\sim p_x} [\sum_{i}^{C} \sum_{h}^{H} \sum_{w}^W y_{i}(h, w) \log \left(G(x_{i} (h, w))\right)]
\label{eq:gce}
\end{equation}

\noindent where $p_{data}$ is the original data distribution, $p_x$ is the data distribution of the artificially corrupted patches, $C=\{0,1\}$ is the set of classes (binary segmentation), $W$ and $H$ are the width and the height of the patch respectively . While other image inpainting approaches typically use $\ell_1$ or $\ell_2$ regression losses, we found that $\mathcal{L}_{CE}$ converges better with less artifacts.

\subsection{Local Discriminator $D_L$}
\label{sec:ld}
While we do have supervision, we have found that pure supervision leads to lack of preservation even of local structure and topology. We learn a data-driven loss via a local discriminator $D_L$ (a fully convolutional network) which takes a patch as input and classifies from which distribution the input comes from (e.g. complete or inpainted patch). 
To make $D_L$ more specific to local high-frequency details and thus be a better data-driven loss, we adopt a Markovian discriminator as in PatchGAN \citep{Isola2017ImagetoImageTW}. 
$D_L$ operates convolutionally over the inputs. For example, given an  image of size $256\times 256$, it is divided into small local patches of size $128 \times 128$. These local patches are provided to $D_L$ sequentially to obtain a decision for each patch. Decision scores are then averaged together to obtain the final prediction for that inputs. 
$D_L$ is optimized using the LSGAN loss \cite{Mao2017} as follows:
\begin{equation}
\mathcal{L}_{LocDadv} = \mathbb{E}_{y\sim p_{data}}[(1 - D_L(y))^2]  + \mathbb{E}_{x \sim p_x}[D_L(G(x))^2],
\label{eq:localdadv}
\end{equation}
\begin{equation}
\label{eq:localgadv}
\mathcal{L}_{LocGadv}= \mathbb{E}_{x \sim p_{x}}[(1 - D_L(G(x)))^2].
\end{equation}
We observe more stable training (fewer vanishing gradients) and reduced mode collapse with LSGAN in agreement with \cite{Mao2017}. We further adopt Spectral Normalization \cite{Miyato2018SpectralNF} to train $D_L$ to constraint its Lipschitz constant, thus smoothing the curvature of the loss landscape further improving training stability.  

\subsection{Global Discriminator $D_G$}
When $G$ makes predictions on input large-scale images, it may produce results of high local inpainting quality, but lack global completeness, as demonstrated in \Cref{fig:compare-gan-gl}. To address this problem without saturating the GPU memory, as we deal with large-scale images, we devise a global discriminator $D_G$, to learn a data-driven loss that aims to capture global structure and topology. The design and training of $D_G$ is a key contribution of this work.

\begin{figure}
    \centering
    \includegraphics[width=0.9\linewidth]{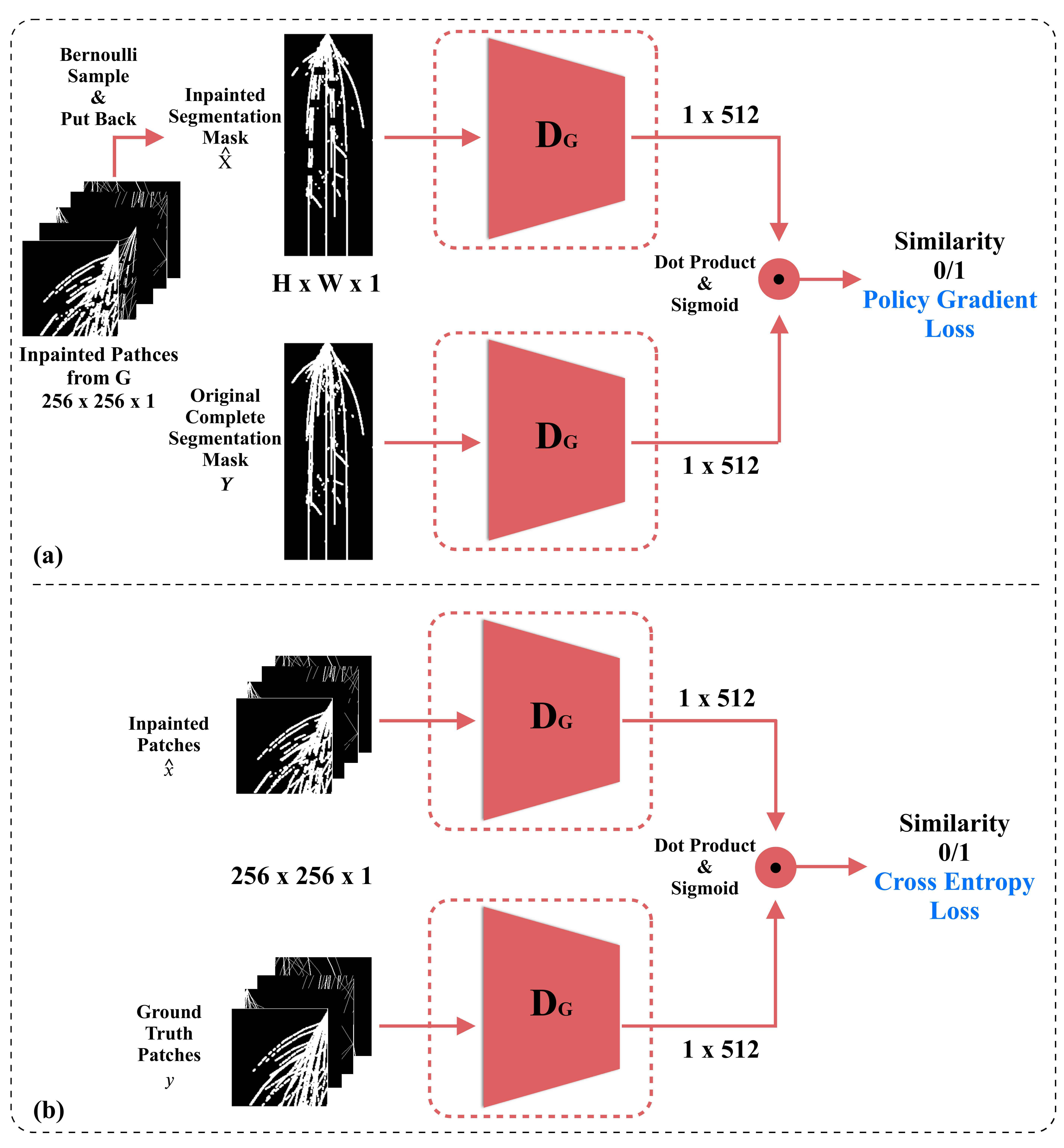}
    \caption{Operation of the global discriminator $D_G$. (a) $D_G$ takes $Y$ and an inpainted $\hat{X}$ large-scale segmentation masks, to provide gradient estimated via Policy Gradient (\cref{eq:globalgadv}) to update the weights of the generator $G$; (b) $D_G$ takes patches as inputs when updating its own weights with gradients computed via the cross entropy loss (\cref{eq:gdadv}) function.}
    \label{fig:fig6-DG}
\end{figure}

$D_G$ (architecture detailed in the supplemental) takes two segmentation masks as inputs: $Y$ the large-scale complete segmentation examples from which the patches used for training the generator were extracted (before gaps were introduced) and $\hat{X}$ the inpainted output after patches have been inpainted and re-composed into an image. Its purpose is to estimate a similarity between $Y$ and $\hat{X}$. 

$Y$ and $\hat{X}$ are processed sequentially and are mapped into two feature vectors (each of dimension 512) via a fully convolutional feature extractor $f$. The similarity of these two vectors is then computed via a dot product and passed via a sigmoid activation to obtain the similarity (reward, $r$) score of these two inputs. We formulate this process as:
\begin{equation}
\label{eq:gdsimilarity}
D_G(Y, \hat{X}) = r(Y, \hat{X}) = \sigma(f(Y), f(\hat{X})).
\end{equation}

The generator $G$ can interact with $D_G$ adversarially by generating inpainting outputs that make $\hat{X}$ more similar to $Y$. The training process and how $D_G$ interacts with $G$ is illustrated in \cref{fig:fig6-DG}. To make the model patch-based, we use the inpainted patches to reconstruct $\hat{X}$. To avoid the backpropagation on large-scale segmentation masks, we still use patches to update the weights of $D_G$ via an adversarial loss (defined in \cref{eq:gdadv} below). In contrast, the weights of $G$ are updated through a policy gradient loss using the REINFORCE procedure \citep{williams1992reinforcement}) (detailed in \cref{eq:globalgadv} below).  

When updating the generator $G$ through $D_G$, we view our generator $G$ as a policy network. It takes corrupted patches as input, and produces probability maps, indicating the probability of each element being an object pixel, which could be viewed as actions taken according to the states. A Bernoulli sampling is then conducted on the probability maps to obtain binary inpainting results. Compared to plain thresholding the output, Bernoulli sampling further improves the inpainting confidence in a probabilistic fashion. Afterwards, the inpainted patches are re-composed in the corrupted large-scale segmentation to obtain a large-scale inpainted mask.  

The global discriminator $D_G$, ie., \cref{eq:gdsimilarity}, is viewed as our value function. We train $D_G$ with the cross entropy loss \cite{Goodfellow2014GenerativeAN}, since the last layer of $D_G$ is a sigmoid, and $G$ with the policy gradient loss we reviewed in \Cref{sec:pg}:
\begin{equation}
\label{eq:gdadv}
\begin{split}
\mathcal{L}_{GloDadv} &= \mathbb{E}_{y_1, y_2 \sim p_{y}}[\log(D_G(f(y_1), f(y_2)))] \\
&+ \mathbb{E}_{{x_1} \sim p_{x}}[\log(1 - D_G(f(G(x_1)), f(y_1))],
\end{split}
\end{equation}
\begin{equation}
\label{eq:globalgadv}
L_{GloGadv}= \mathbb{E}_{\hat{X} \sim p_{\hat{X}}, Y \sim p_{Y}, x \sim p_{x}}[D_G(f(\hat{X}), f(Y)) \log(G(x))],
\end{equation}
where both $y_1$ and $y_2$ are complete patches but extracted from different segmentation masks, and $\hat{X}$ is the large-scale images where inpainted patches have been relocated back in the corrupted mask $X$. As previously described in \cref{sec:ld}, we also adopt Spectral Normalisation \cite{Miyato2018SpectralNF} for $D_G$.

\subsection{Combined costs: the GAN-GL Model}
The training loss function for the generator is given by the combination of \cref{eq:gce,eq:localgadv,eq:globalgadv}:
\begin{equation}
\label{eq:totalgenerator}
\mathcal{L}_G = \lambda_1  \mathcal{L}_{CE} + \lambda_2 \mathcal{L}_{LocGadv} + \lambda_3 \mathcal{L}_{GloGadv},
\end{equation}
where $\lambda_1$, $\lambda_2$ and $\lambda_3$ balance the influence of each term.  The local discriminator is updated w.r.t. \cref{eq:localdadv}, whereas the global discriminator is updated w.r.t. \cref{eq:gdadv}, both using patches.

\subsection{A Masked Loss Variant: the GAN-GL-M Model}
\label{sec:ganglm}
In real input images, such as the images of roots, may already contain inherent gaps, but we do not want the model to learn this data bias. Here, we propose a variant (referred as \textbf{GAN-GL-M}) that uses the position of the artificial gaps in the loss (and not as an input). When computing the gradients of the generator, the loss is computed only in the locations of the artificially introduced gaps, using a masked loss  \cite{wu2017maskedloss}:
%
\begin{equation} 
\mathcal{L}_{MCE} = L_{CE}(y \odot M, \hat{x} \odot M)
\label{eq:gmaskce}
\end{equation}
\noindent where $M$ indicates the positions of the artificial gaps, and $\odot$ is the element-wise multiplication of tensors.  Note that this loss can be used in lieu of the \cref{eq:gce} of $G$'s objective when needed at the batch level as we detail below.

\subsection{Training and Inference Details}
\label{sec:trainingdetails}

Algorithm~\ref{alg:training} lists training pseudo-code, split into $3$ parts: (i) update local discriminator $D_L$; (ii) update global discriminator $D_G$; (iii) update the inpainting generator $G$. Both (i) and (ii) are conducted on patch-level of size $256 \times 256$, whereas (iii) is conducted on both patch-level and global-level of various sizes.  GAN-GL is trained using Algorithm~\ref{alg:training}. A batch size of $8$ is adopted and patches within one batch are come from the same large-scale mask. To train on mix of synthetic and real data (with inherent gaps) using GAM-GL-M, we alternate two different batch settings: one batch updates both generator and discriminators using Algorithm~\ref{alg:training} on synthetic data; the other batch uses \cref{eq:gmaskce} to train only the generator on real data. We adopt different learning rates for generator and the discriminator \citep{heusel2017gans}, which improves the convergence of GANs. 
At inference time, we only need $G$, which takes few milliseconds per patch of size $256 \times 256$.

\begin{algorithm}[!t]
	\begin{algorithmic}[1]
		\FOR{number of total training iterations}
		\STATE Sample a batch of $m$ complete patches $y_1 = \{y_1^{(1)}, y_1^{(2)}, ..., y_1^{(m)}\}$ from large-scale mask $Y_1$
		\STATE Sample a batch of $m$ complete patches $y_2$ from another large-scale mask $Y_2$
		\STATE Introduce gaps into complete patches $y_1$ to make corrupted patches $x_1 = \{x_1^{(1)}, x_1^{(2)}, ..., x_1^{(m)}\}$
		\STATE Forward through $G$ to obtain probability maps $\hat{x}_1 = \{
		\hat{x}_1^{(1)}, \hat{x}_1^{(2)}, ..., \hat{x}_1^{(m)}\}$
		\STATE Update $D_L$ with thresholded $\hat{x}_1$ and $y_1$: \\
		\begin{center}
			\vskip 1mm
			$\nabla_{\theta_{D_L}} \frac{1}{m} \sum_{i=1}^m \big [ (1 - D_L(y_1^{(i)}))^2 + D_L(G(\hat{x}_1^{(i)}))^2 \big]$
			\vskip 1mm
		\end{center}
		\STATE Update $D_G$ with thresholded $\hat{x}_1$, $y_1$ and $y_2$: \\
		\begin{center}
			\vskip 1mm
			$\nabla_{\theta_{D_G}} \frac{1}{m} \sum_{i=1}^m \big [ \log{r(f(y_1^{(i)}), f(y_2^{(i)}))} + [\log{(1 - r(f(\hat{x}_1^{(i)}), f(y_1^{(i)})))} \big]$
			\vskip 1mm
		\end{center}
		\STATE Bernoulli sample on $\hat{x}_1$ and put them back to compose $\hat{X}_1$
		\STATE Update $G$ with $\hat{x}_1$, $y_1$, $\hat{X}_1$ and $Y_1$: \\
		\begin{center}
			\vskip 1mm
			$\nabla_{\theta_{G}} \frac{1}{m} \sum_{i=1}^m \big [ 
			\lambda_1\big(\log{G(\hat{x}_1^{(i)})} + (1-\log(G(\hat{x}_1^{(i)})) \big) + 
			\lambda_2 \big( (1 - D_L(G(\hat{x}_1^{(i)})))^2 \big) +
			\lambda_3 \big( r(f(\hat{X}_1), f(Y_1)) \log(\hat{x}_1^{(i)}) \big)
			\big]$
			\vskip 1mm
		\end{center}
		\ENDFOR
	\end{algorithmic}
	\caption{Training procedure of our proposed model}
	\label{alg:training}
\end{algorithm}

\section{Experimental Results}
\label{sec:er}


\subsection{Experimental Setup}
\mypar{Datasets} We train and evaluate our model on $4$ datasets.\\
(i) \textbf{Roots} \cite{lobet_guillaume_2016_61739, Bontpart573139}. This set has segmentation masks from synthetic \cite{lobet_guillaume_2016_61739} and real roots \cite{Bontpart573139}. Each real root mask has dimension of approximately $2500 \times 1000$ and does not have ground truth. For all root experiments, except the plant study in \Cref{sec:root-ablation}, we train our models on only synthetic root but evaluate them on synthetic and root. Our synthetic root data are augmented to transform their appearance to resemble real roots, by skeletonizing, adding noise, dilation, and rotation.\\
(ii) \textbf{HRF retina} \cite{odstrcilik2013retinal}. This dataset has $45$ retinal vessel segmentation masks of size $3504 \times 2336$. Due to the relatively small size of the HRF dataset, we select segmentation masks that have only one fully connected component for training and use masks that are incomplete for qualitative evaluation only.\\
(iii) \textbf{Road Detection} \citep{MnihThesis}. This dataset has around $1,100$ images of roads of size $1488 \times 1488$. \\
(iv)  \textbf{Line sketches}  by Leonardo Da Vinci \cite{sasaki2018learning}.  This dataset contains images that have various sizes but smaller compared to the other datasets. 

\begin{table*}[!h]
\centering
\caption{Comparison of patch inpainting results. Bold fond indicates statistical significance of GAN-GL over the baselines ($p <0.0001$)}
\label{tab:patch-comparison}
\resizebox{0.9\textwidth}{!}{%
\begin{tabular}{@{}c|ccccc@{}}
\toprule
Datasets & Models & MSE Overall $\downarrow$ & MSE Within Gaps $\downarrow$ & Relative Pixel $\uparrow$ Diff. & Relative Comp. Diff. $\uparrow$ \\ \midrule
\multirow{3}{*}{\begin{tabular}[c]{@{}c@{}}Synthetic\\ Root\\ ($N=1,500$)\end{tabular}} & CNN & .0065 (.0043) & .3832 (.1748) & .6257 (.1748) & .8033 (.2807) \\
 & U-Net & .0080 (.0046) & .3743 (.1684) & .6329 (.1684) & .8726 (.2223) \\
 & GAN-GL & \textbf{.0036 (.0031)} & \textbf{.2472 (.1498)} & \textbf{.6986 (.1498)} & \textbf{.9358 (.1744)} \\ \midrule
\multirow{3}{*}{\begin{tabular}[c]{@{}c@{}}Real\\ Root\\ ($N=400$)\end{tabular}} & CNN & .0114 (.0056) & .4309 (.1245) & .5691 (.1245) & .8032 (.2026) \\
 & U-Net & .0131 (.0059) & .4379 (.1338) & .5181 (.1415) & .8114 (.2126) \\
 & GAN-GL & \textbf{.0087 (.0048)} & \textbf{.2598 (.1196)} & \textbf{.6941 (.1348)} & \textbf{.9250 (.1195)} \\ \midrule
\multirow{3}{*}{\begin{tabular}[c]{@{}c@{}}Retinal\\ Vessel\\ ($N=180$)\end{tabular}} & CNN & .0119 (.0072) & .4463 (.2100) & .5537 (.2100) & .6549 (.2558) \\
 & U-Net & .0138 (.0086) & .4994 (.2089) & .5006 (.2089) & .6577 (.2625) \\
 & GAN-GL & \textbf{.0099 (.0064)} & \textbf{.3800 (.2058)} & \textbf{.6200 (.2058)} & \textbf{.7499 (.2355)} \\ \midrule
\multirow{3}{*}{\begin{tabular}[c]{@{}c@{}}Road\\ ($N=210$)\end{tabular}} & CNN & .0005 (.0005) & .0472 (.0859) & .9528 (.0859) & .9884 (.0771) \\
 & U-Net & .0005 (.0006) & .0447 (.0761) & .9553 (.0761) & .9847 (.0793) \\
 & GAN-GL & \textbf{.0003 (.0004)} & \textbf{.0326 (.0671)} & \textbf{.9674 (.0671)} & .9912 (.0691) \\ \midrule
\multirow{3}{*}{\begin{tabular}[c]{@{}c@{}}Line\\ Sketch\\ ($N=60$)\end{tabular}} & CNN & .0019 (.0010) & .2106 (.1085) & .7894 (.1085) & .8820 (.1275) \\
 & U-Net & .0016 (.0009) & .1823 (.1002) & .8179 (.1002) & .8889 (.1459) \\
 & GAN-GL & .0016 (.0009) & .1868 (.1034) & .8130 (.1034) & \textbf{.8960 (.1405)} \\ \bottomrule
\end{tabular}%
}
\end{table*}

\mypar{Evaluation metrics} We evaluate our model on  patch- and image-level. For patch-level evaluation, as real root segmentation and retinal segmentation masks contain inherent gaps, we visually select a set of patches from the held out test set that contains no gaps. We use mean squared error (MSE) and pixel difference as metrics for our experiments artificially corrupted inputs and the inpainted results. Furthermore, we use difference of number of fully connected components to measure the completeness of the inpainting results as proxy metric. We normalize pixel difference and fully connected components difference, using the relative improvement obtaining after inpainting $\frac{|metric - \overline{metric}|}{metric}$,
where $metric$ is the metric value before inpainting, and $\overline{metric}$ is the same metric calculated after inpainting. We repeat experiments $3$ times using different random initialization and report average results. 


\mypar{Baselines and ablated models} We compare our method (\textbf{GAN-GL}) with $2$ baselines, namely the \textbf{CNN} proposed in \cite{Sasaki2017JointGD} and the \textbf{U-Net} architecture in \cite{Chen2018RootGC}. 

In our study on roots in \Cref{sec:root-ablation}, we also consider using \textit{only} the local discriminator $D_L$ (referred as \textbf{GAN-L}) and using \textit{only} the global discriminator $D_G$ (referred as \textbf{GAN-G}) to showcase the importance of the global discriminator. 

To bridge the domain gap between synthetic and real data,\footnote{Synthetic data are simpler in structure and do not contain root hairs that in the real data appear as 1-pixel-wide texture around the roots.} we employ the \textbf{GAN-GL-M} variant with masked loss to train combining synthetic and root data. 

\mypar{Hyper-parameters} Experiments on roots use $2\times 10^{-4}$ as learning rate for $G$ and $4\times 10^{-4}$ as learning rates for the discriminators.  
Remaining experiments, used reduced learning rates by a factor of two.  We use $\lambda_1=1000$, $\lambda_2=\lambda_3=1$ in GAN-GL (and GAN-GL-M) for all experiments.


\begin{table*}[]
\centering
\caption{Comparison of whole segmentation mask inpainting results. Bold fond same meaning as in \cref{tab:patch-comparison}. } 
\label{tab:whole-comparison}
\resizebox{0.9\textwidth}{!}{%
\begin{tabular}{@{}c|ccccc@{}}
\toprule
Datasets & Models & MSE Overall $\downarrow$ & MSE within Gaps $\downarrow$ & Relative Pixel Diff. $\uparrow$ & Relative Comp Diff. $\uparrow$ \\ \midrule
\multirow{3}{*}{\begin{tabular}[c]{@{}c@{}}Synthetic \\ Root\\ ($N=400$)\end{tabular}} & CNN & .0044 (.0044) & .0012 (.0013) & .7598 (.3258) & .8080 (.4748) \\
 & U-Net & .0021 (.0027) & .0011 (.0012) & .8262 (.3613) & .8635 (.4305) \\
 & GAN-GL & \textbf{.0008 (.0009)} & \textbf{.0007 (.0009)} & \textbf{.9346 (.2156)} & \textbf{.9568 (.1625)} \\ \midrule
 \multirow{3}{*}{\begin{tabular}[c]{@{}c@{}}Road\\ ($N=48$)\end{tabular}} & CNN & .0092 (.0020)) & .0086 (.0017) & .0688 (.0401) & .9568.(.0531) \\
 & U-Net & .0088 (.0015) & .0087 (.0015) & .0975 (.0130) & .9535 (.0226) \\
 & GAN-GL & \textbf{.0084 (.0013)} & \textbf{.0082 (.0012)} & \textbf{.1120 (.0086)} & \textbf{.9866  (.0141)} \\ \midrule
\multirow{3}{*}{\begin{tabular}[c]{@{}c@{}}Line\\ Sketch\\ ($N=8$)\end{tabular}} & CNN & .0506 (.0006) & .0354 (.0007) & .2467 (.0168) & .8543 (.3521) \\
 & U-Net & .0352 (.0020) & .0334 (.0014) & .2653 (.0165) & .8564 (.3342) \\
 & GAN-GL & \textbf{.0326 (.0010)} & \textbf{.0310 (.0010)} & \textbf{.2987 (.0146)} & \textbf{.8813 (.2980)} \\ \bottomrule
\end{tabular}%
}
\end{table*}

\begin{table*}[]
\centering
\caption{Ablation study on plant root patches. A * indicates statistical significance of GAN-GL over both GAN-G and GAN-L with $p$-value $<0.0001$. The bold fond indicates statistical significance of GAN-GL-M over GAN-GL.}
\label{tab:patch-ablation}
\resizebox{0.9\textwidth}{!}{%
\begin{tabular}{@{}c|ccccc@{}}
\toprule
Datasets & Models & MSE Overall $\downarrow$ & MSE Within Gaps $\downarrow$ & Relative Pixel Diff. $\uparrow$ & Relative Comp. Diff. $\uparrow$ \\ \midrule
\multicolumn{1}{c|}{\multirow{4}{*}{\begin{tabular}[c]{@{}c@{}}Synthetic\\ Root\\ ($N=1,500$)\end{tabular}}} & GAN-L & .0039 (.0034) & .2762 (.1693) & .6709 (.1693) & .9011 (.2031) \\
\multicolumn{1}{c|}{} & GAN-G & .0041 (.0041) & .2631 (.1580) & .6529 (.1580) & .9018 (.2143) \\
\multicolumn{1}{c|}{} & GAN-GL & .0036 (.0031)* & .2472 (.1498)* & .6986 (.1498)* & .9358 (.1744)* \\
\multicolumn{1}{c|}{} & GAN-GL-M & \textbf{.0026 (.0020)} & \textbf{.2360 (.1450)} & \textbf{.7540 (.1450)} & .9360 (.1767) \\ \midrule
\multicolumn{1}{c|}{\multirow{4}{*}{\begin{tabular}[c]{@{}c@{}}Real\\ Root\\ ($N=400$)\end{tabular}}} & GAN-L & .0093 (.0050) & .2794 (.1173) & .6636 (.1428) & .9024 (.1323) \\
\multicolumn{1}{c|}{} & GAN-G & .0092 (.0052) & .2711 (.1516) & .6489 (.1516) & .8985 (.1454) \\
\multicolumn{1}{c|}{} & GAN-GL & .0087 (.0048)* & .2598 (.1196)* & .6941 (.1196)* & .9250 (.1195)* \\
\multicolumn{1}{c|}{} & GAN-GL-M & \textbf{.0061 (.0039)} & \textbf{.2260 (.1157)} & \textbf{.7639 (.1157)} & \textbf{.9416 (.1174)} \\ \bottomrule
\end{tabular}%
}
\end{table*}

\begin{figure*}
    \centering
    \includegraphics[width=0.9\textwidth]{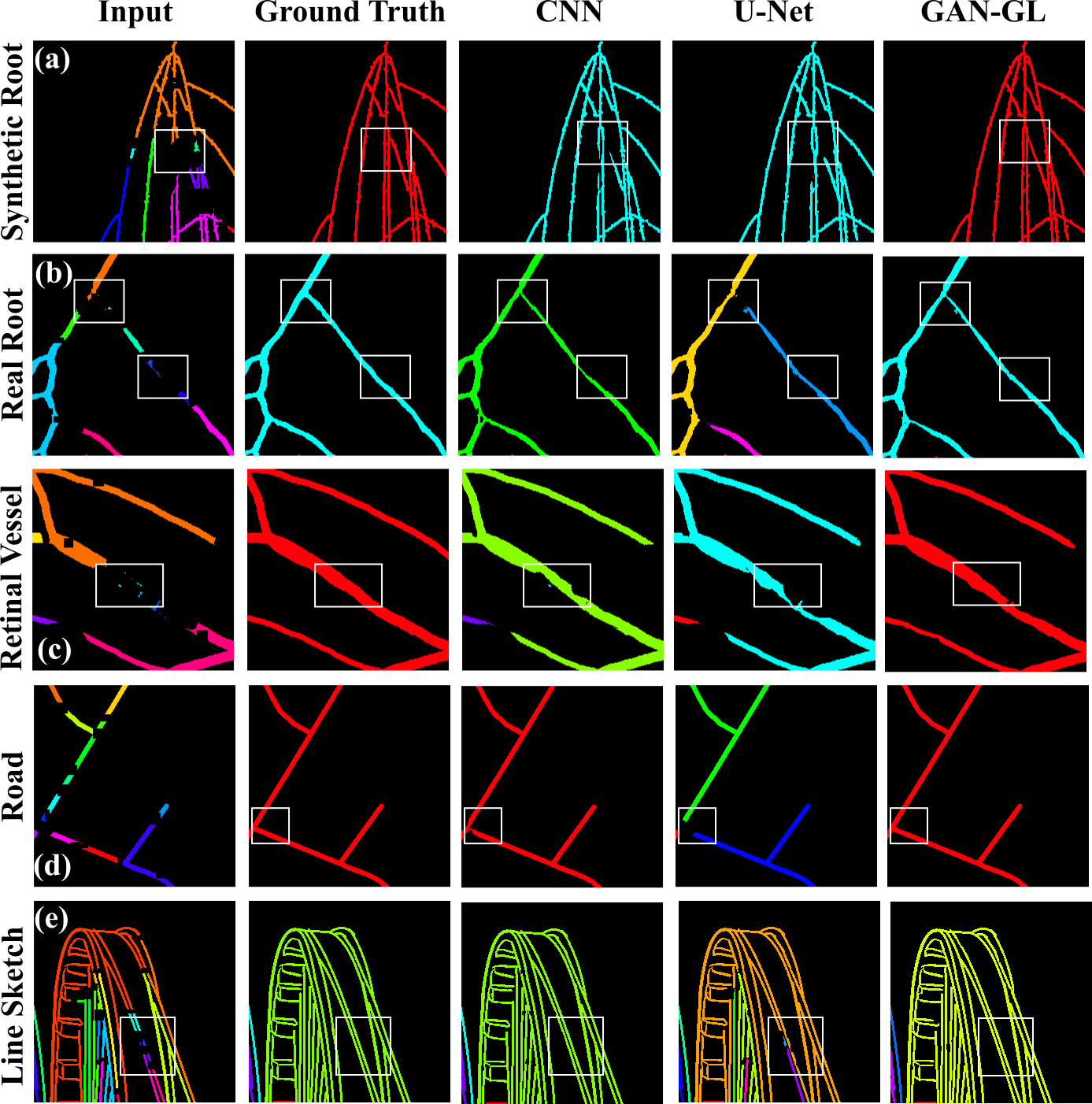}
    \caption{Patch inpainting results of synthetic root (a), real root (b), retinal vessel (c), road (d) and line sketch (e). Different colors indicate segments caused by gaps. GAN-GL produces the most consistent and complete inpainting, with bounding boxes indicating the improvement.}
    \label{fig:fig8-patch-comparison}
\end{figure*}

\subsection{Thin Structure Segmentation Recovery}
\mypar{Patch-level evaluation} For this experiment, we visually select real root and retina vessel segmentation patches that contain no gaps to have ground truth for evaluation purposes only. Results for all datasets at patch-level are shown in \Cref{tab:patch-comparison}. For the root dataset, we highlight that we train using only synthetic root data. 
We performed a paired two-tailed t-test when comparing the results of GAN-GL and the baseline models (CNN \cite{Sasaki2017JointGD} and U-Net \cite{Chen2018RootGC}). 

\begin{figure*}
    \centering
    \includegraphics[width=0.8\textwidth]{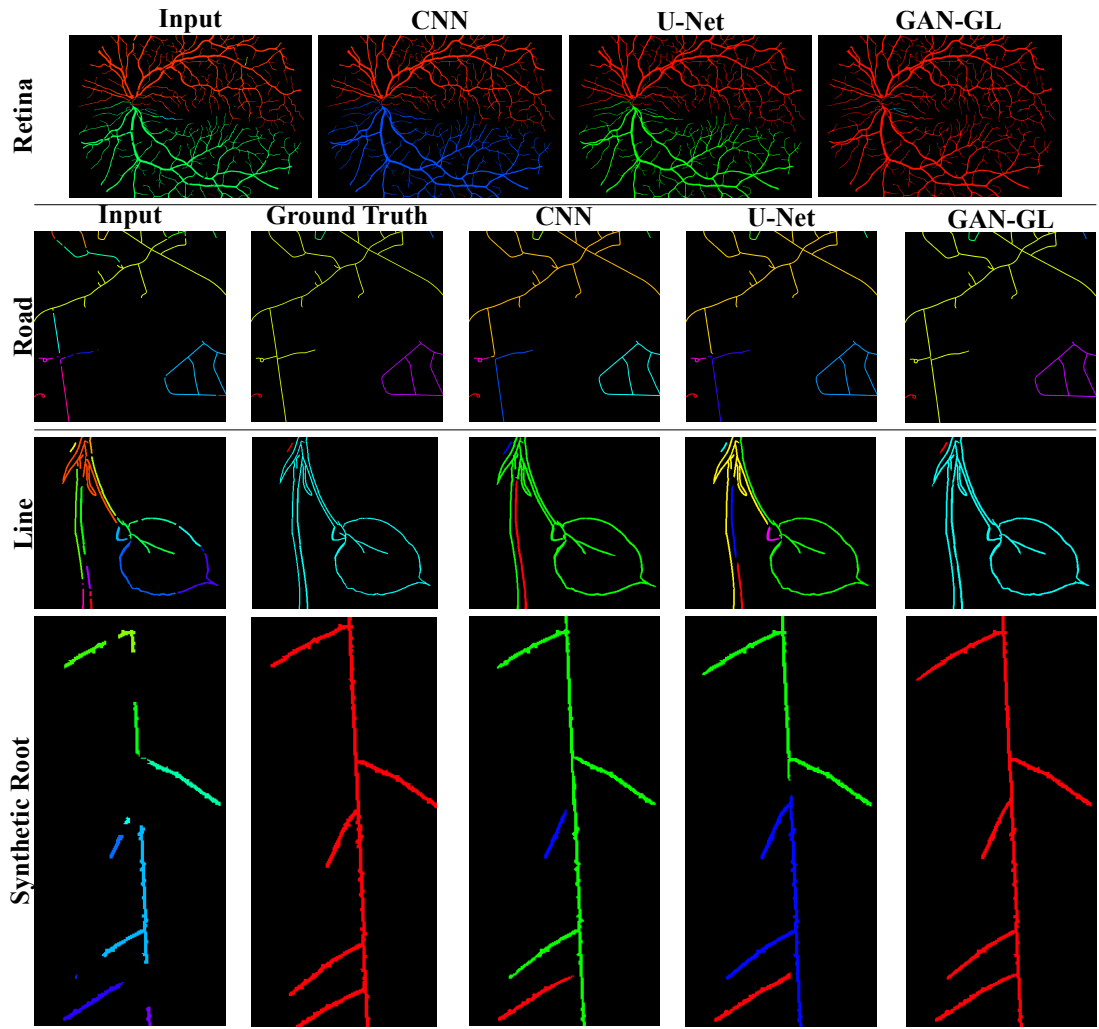}
    \caption{Whole image inpainting results on retinal vessel, satellite, and line drawing sketch binary images. GAN-GL produces the best results.} 
    \label{fig:fig11-all-whole}
\end{figure*}

\mypar{Whole image-level evaluation} We also compare the models at large-scale image-level. We quantitatively  evaluate our models on synthetic roots, road and line drawing sketch datasets,\footnote{As there is no ground truth for whole images in real root and retinal test data we did not include them in this analysis.} which are shown in \cref{tab:whole-comparison}. We also display qualitative results of whole retinal vessel, road, line sketch and synthetic root segmentation masks in \cref{fig:fig11-all-whole}. Further results with the real root inpainting results are shown in  \cref{sec:root-ablation}. 

\vspace{-0.75em}
\mypar{Discussion} In all experiments (both at patch- and  image-level), GAN-GL outperforms other methods. As shown in \cref{tab:patch-comparison} and \cref{tab:whole-comparison}, MSE and pixel difference are significantly improved (statistically significant on most occasions) on all datasets.
For line sketch patches, while our model performs similarly as the U-Net, it still exhibits improvement on completeness, by reducing the number of disconnected components, as it can be seen in \cref{tab:patch-comparison}.   

These observations are more evident qualitatively as the visual examples in \cref{fig:fig8-patch-comparison} show, where root segments resulting from discontinuities are colored differently (fewer number of colors indicates a more complete structural recovery). It is thanks to the global discriminator $D_G$, that our model can produce the most coherent inpainting results. 

Referring to \cref{tab:whole-comparison}   and  \cref{fig:fig11-all-whole},
at image-level our method is able to outperform the other methods also in repairing large-scale segmentation masks across the different datasets considered. GAN-GL produces the most consistent and realistic results, confirming the observations made above. Roads and line sketches have relatively simple structural textures, thus, the differences between the models are less significant and the baseline models could already produce good results due to the simplicity of the dataset. However, statistical significance of our model over the baselines still exists. 

\subsection{A Study on Plant Roots}
\label{sec:root-ablation}
This work commenced for solving the problem of recovering gaps in roots of the chickpea plant, where ground truth is lacking.  While all previous analysis used only synthetic root data, here we show the benefit of training with the combination of synthetic and real data (\cref{sec:trainingdetails}), 
using the masked loss \cref{eq:gmaskce} and the variant GAN-GL-M model (\cref{sec:ganglm}). To showcase the importance of each element in the model, this analysis also serves as an ablation study on using only a local discriminator GAN-L and a model trained with only global discriminator GAN-G.  In addition, since we have argued that gaps affect downstream analysis tasks, we introduce additional metrics that arise in the domain of root analysis (traits to peruse the lingo of plant breeding). 


\begin{table}[t]
\centering
\caption{Comparison of real chickpea whole root results ($N=25$) in terms of relative improvement on the number of fully connected components, root length, tip count, and convex hull area of inpainted results compared to original corrupted ones. A * indicates statistical significance of GAN-GL over baseline models and the bold font indicates  statistical significance of GAN-GL-M over GAN-GL with $p$-value $<0.0001$.}
\resizebox{\linewidth}{!}{%
\begin{tabular}{@{}ccccc@{}}
\toprule
& \begin{tabular}[c]{@{}c@{}}Relative Comp. \\ Diff. $\uparrow$\end{tabular} & Length ($\%$) $\uparrow$ & \# Tips ($\%$) $\uparrow$ & Convex Hull ($\%$) $\uparrow$ \\ \midrule
CNN  & .5637 (.3779) & 6.04 (6.26) & 26.43 (15.79) & 66.39 (13.13) \\
U-Net  & .5882 (.3884) & 7.44 (7.59) & 28.57 (18.42) & 67.61 (13.83) \\
\midrule
GAN-L & .6447 (.4759) & 7.16 (7.08) & 29.29 (18.42) & 70.20 (10.06) \\
GAN-G & .6928 (.4252) & 9.17 (8.85) & 28.57 (14.06) & 68.48 (12.93) \\
GAN-GL & .6854 (.5832)* & 9.63 (10.55)* & 30.00 (19.30)* & 80.48 (22.41)* \\ \midrule
GAN-GL-M & \textbf{.7211 (.5970)} & \textbf{13.58 (14.47)} & \textbf{32.14 (21.93)} & 81.12 (23.94) \\ \midrule
Expert & .7628 (.6739) & 5.71 (5.06) & 39.29 (28.07) & 74.52 (15.26) \\ \bottomrule
\end{tabular}%
}
\label{tab:root-full}
\end{table}

\mypar{Patch-level evaluation} We evaluate GAN-L, GAN-G, GAN-GL, and GAN-GL-M on both synthetic and real root patches.  (Only GAN-GL-M is trained with both synthetic and real root data, all other model variants are trained only with only synthetic data.) The results are shown in \cref{tab:patch-ablation}. 

\begin{figure*}
    \centering
    \includegraphics[width=\textwidth]{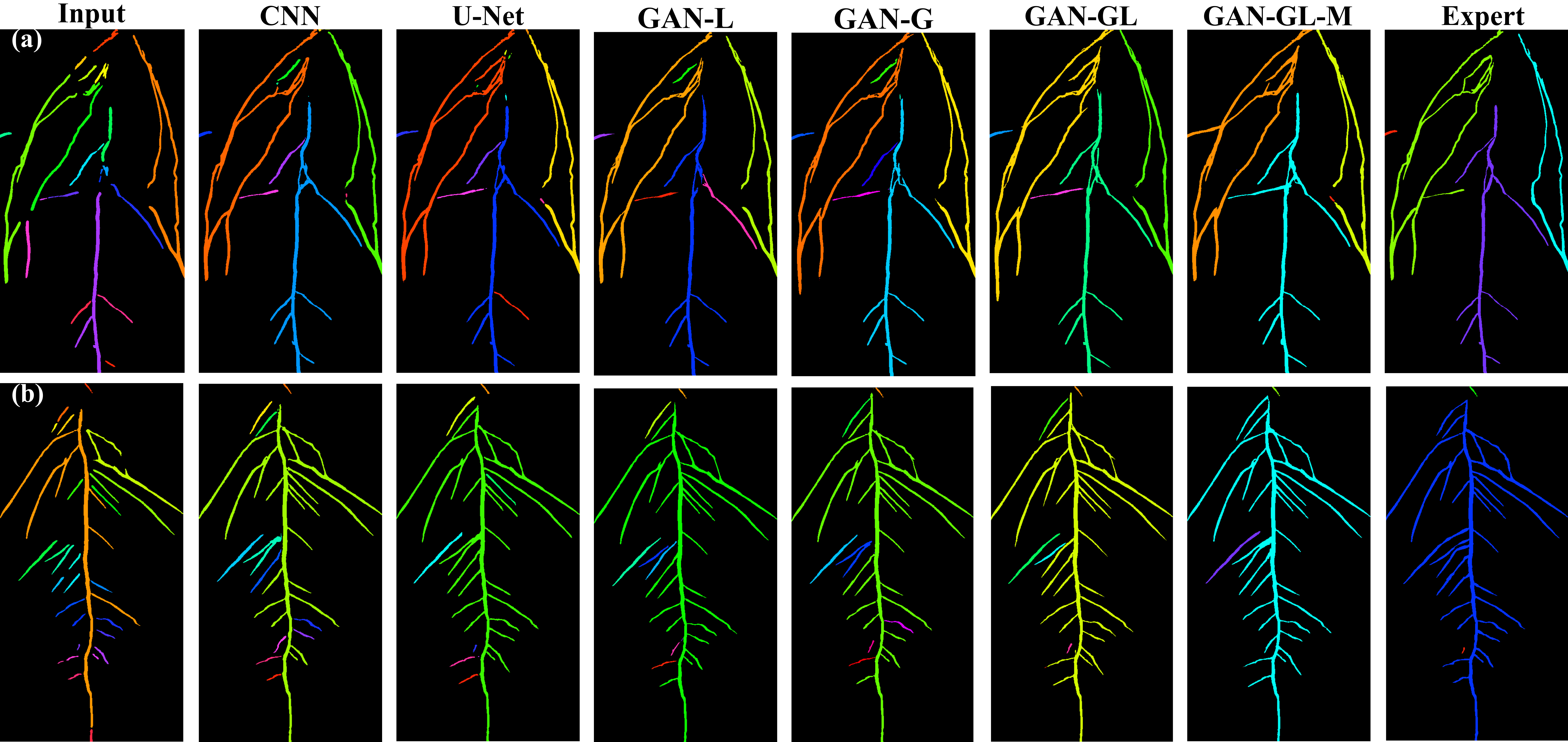}
    \caption{Inpainting results on whole chickpea root (a)-(b) from different models. GAN-GL-C produces inpainting results that are similar to expert-level.}
    \label{fig:fig9-whole-root}
\end{figure*}

Including a local discriminator (GAN-L), the performance already improves upon the U-Net baseline (results in \cref{tab:patch-comparison}) \cite{Chen2018RootGC}. 
A local discriminator helps to inpaint more accurately, yet some gaps are still left incomplete (see \cref{fig:fig9-whole-root} for visual evidence). The global discriminator alone (GAN-G) does not seem to offer any considerable improvement numerically and visually. With the two discriminators combined, GAN-GL  offers significant improvement. GAN-GL-M further improves this, as now the model is trained  with a combination of real and synthetic data. The scores are considerably improved on both cases. It also further boosts the performance of the model on connecting the segments in real root patches.

\mypar{Trait evaluation of whole segmentation masks and comparison with expert annotators} To appreciate the impact of this improvement in down-stream applications, we used the Root Image Analysis-J (RIA-J) software \citep{Lobet17} to extract several root traits from whole root segmentations, such as root length, tip counts, and convex hull area. These metrics are measured on real root segmentations (that have gaps)  and their inpainted outputs, with relative improvement computed as previously defined. Since no ground truth exists in real root data, an expert annotated $25$ full roots, hence providing a human-level correction of the gaps as a reference. We show quantitative results in \Cref{tab:root-full} and qualitative examples in \Cref{fig:fig9-whole-root}.


Our proposed approach (both GAN-GL and GAN-GL-M) can produce inpainting results that have a lower number of fully connected components (an indication of being more complete), which is statistically significant over the results of single discriminator models (GAN-G and GAN-L). The real root images exhibited several discontinuities, resulting in inaccurate tip counts, which our model `repairs' effectively. Also, inpainted results from our model have larger convex hull area, indicating more discontinuities are reconnected together. One can also observe that GAN-GL-M achieves human-level performance (compare with  `Expert' in \cref{tab:root-full} and \cref{fig:fig9-whole-root}). While the measurements of GAN-GL-M show improved root length and convex hull area, the expert annotations show less improvement on these two traits, indicating that maybe our model is more consistent. 

\section{Conclusion}
\label{sec:conclusion}

We present an effective approach for inpainting gaps in segmentation masks of thin structures  in an adversarially way. We use both a local discriminator to encourage local high-quality inpainting results and a global discriminator to encourage global coherency. Our model can be viewed as a general post-processing technique for any binary image input.  Our model could be trained on patch-level but still produce inpainting results that are consistent within large-scale segmentation masks through the interaction between the generator and the global discriminator via the Policy Gradient procedure. We show the generalization of our methods to several scenarios, where our results outperform the state-of-the-art results on $4$ thin structure segmentation datasets. With the analysis on root dataset, we demonstrate that by applying our model on image segmentation masks, the accuracy of the root characteristics extracted from the inpainted images has been improved considerably. Also, our model can outperform human expert on root datasets by providing more natural, realistic, and fine-grained inpainting details. The proposed method has great potential for many real-world applications.


\section*{Acknowledgements}
This work was supported by the BBSRC grant BB/P023487/1 (\url{http://chickpearoots.org}) and also partially supported by The Alan Turing Institute under the EPSRC grant EP/N510129/1. We thank NVIDIA Inc. for providing us with a Titan Xp GPU used for our experiments.

\bibliographystyle{unsrt}
\bibliography{ref} 

\end{document}


\title{Supplementary Material\vspace{-2em}}

\maketitle

\IEEEdisplaynontitleabstractindextext

\vspace{-20cm}
\section{Model Architecture}
\label{sec:arch}

\subsection{Generator $G$}

\begin{figure}
    \centering
    \includegraphics[width=\linewidth]{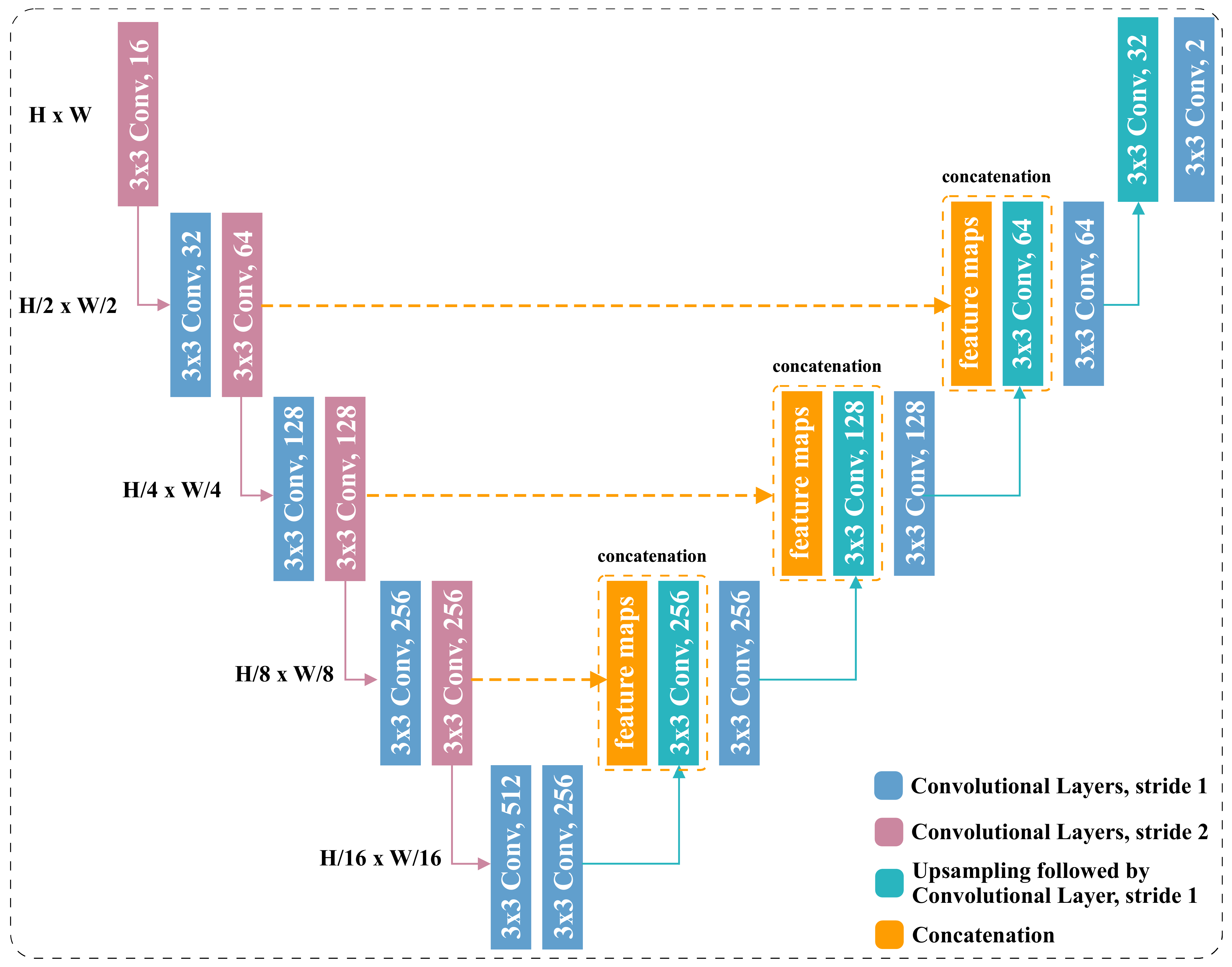}
    \caption{Inpainting generator $G$ architecture. Each convolutional layer is followed by a \textit{Batch Normalisation} layer and a ReLU activation, except the last one, whose feature maps go directly into a Softmax activation to obtain probability maps. All convolutional layers use $3 \times 3$ kernels.}
    \label{fig:fig1-generator-architecture}
\end{figure}

The inpainting generator $G$ is a fully convolutional network \citep{Shelhamer2017fcn} which has an encoder $G_E$ and decoder $G_D$. 

The detailed architecture of the inpainting generator $G$ is shown in \cref{fig:fig1-generator-architecture}. $G$ contains convolutional layers which have a kernel size of $3 \times 3$. Zero-padding is adopted for all the convolutional layers to ensure the images do not shrink during processing. The encoder part of the generator $G_E$ contains $4$ down-sampling operations using convolutional layers of a stride 2, which allows the model to compute features and recognize structure from a larger receptive field. In the decoder path, where the feature maps are expanded, nearest neighbour up-sampling operations are used followed by convolutional layers, which are demonstrated that they could produce better inpainting results than de-convolutional layers \cite{Sasaki2017JointGD}. Due to the aggressive down-sampling operations which compress inputs images by $4$ times smaller, lots of useful features might be lost during the process. Herein, we use the skip connections \cite{RFB15unet} to preserve the features learned at shallower layers and allow the decoder to use them at corresponding positions for the final reconstruction. The information sharing between the encoder and the decoder can improve the model's performance in terms of overall reconstruction accuracy \cite{Chen2018RootGC}. These skip connections also enable a faster and more stable training process as they provide better gradient flow. Each convolutional layer is followed by a Batch Normalisation (BN) \cite{ioffe2015bn} layer that accelerates the training process and a Rectified Linear Unit (ReLU) activation, except the last one. The last layer of the decoder is a convolutional layer producing a feature map with two channels. A softmax activation is then applied on the final feature map to obtain a probability map.

\subsection{Local discriminator $D_L$}

$D_L$ is composed of 6 convolutional blocks. For the first $5$ blocks, each of them consists of a convolutional layer, which has a kernel size of $3$ and a stride of $2$, a spectral normalisation layer \cite{Miyato2018SpectralNF}, a batch normalisation layer \cite{ioffe2015bn} and a Leaky ReLU activation. The spectral normalisation is used to constraint the Lipschitz constant of the classification function defined by $D_L$, which ease the instability of the training process. The last block consists of a convolutional layer which has a kernel size of 4 and a stride of 1 and a sigmoid activation, mapping the feature maps from the last layer into a scalar classification score. We adopt the least square adversarial loss in LSGAN \cite{Mao2017} as it provides higher quality results and more stable training process compared to vanilla GAN. \cite{Goodfellow2014GenerativeAN}. 

\begin{figure}
    \centering
    \includegraphics[width=\linewidth]{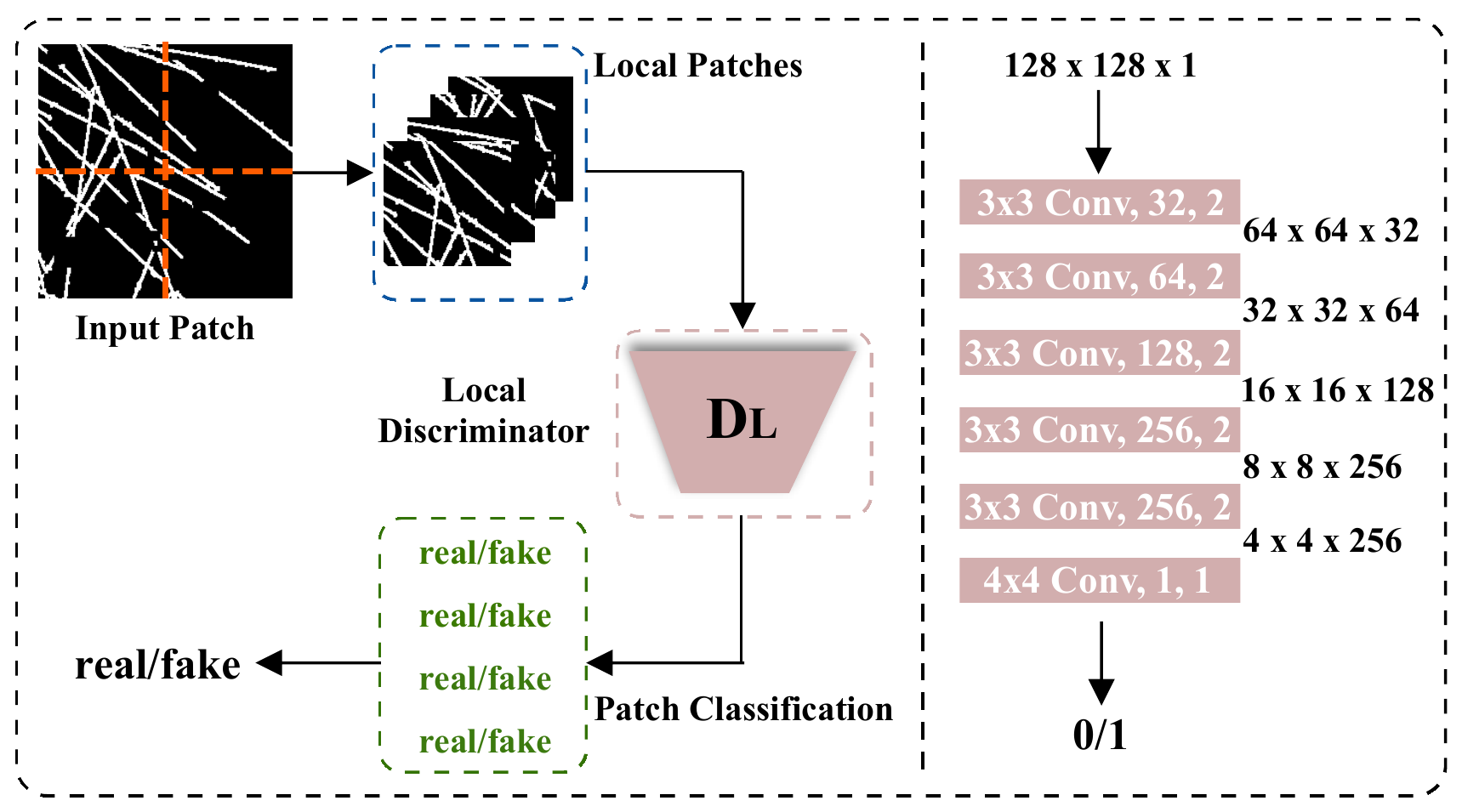}
    \caption{Local discriminator $D_L$ architecture. Each convolution is followed by a \textit{Spectral Normalisation} layer, a \textit{Batch Normalisation} layer, and a Leaky ReLU activation.}
    \label{fig:fig2-DL}
\end{figure}

\subsection{Global Discriminator $D_G$}

The architecture of the global discriminator consists of a series of convolutional layers, an average pooling layer, and a dot product and a sigmoid unit that is used to compute the similarity score. The convolutional architecture in $D_G$ is similar to that of $D_L$, where the last convolutional layer is replaced with another convolutional layer of stride of 2 and all layers have kernel size of $5$. An adaptive average pooling layers is employed to enable $D_G$ to deal with images of different sizes and convert features maps from the last convolutional layer into feature vectors. A dot product between the feature vectors is computed. Then a sigmoid activation maps the results from the dot product into $[0, 1]$ as the final similarity score. \todo{this paragraph (and maybe fig 5) should go to a supplemental material for the details of the architecture(s)}

\bibliographystyle{unsrt}
\bibliography{ref}